\documentclass[11pt]{article}
\usepackage[utf8]{inputenc}
\usepackage[greek,swedish,english]{babel}
\usepackage{alphabeta}
\usepackage{makecell}
\usepackage{chngpage}
\usepackage{adjustbox}
\usepackage{graphicx}
\usepackage{cite}
\usepackage{pgfplots}
\usepackage{amsmath}
\usepackage{caption}
\title{\textbf{A Computational Analysis of Natural Languages to Build a Sentence Structure Aware Artificial Neural Network}}
\author{Alberto Calderone, Ph.D.\\
		sinnefa@gmail.com}
\date{\today}
\begin{document}
\maketitle
\begin{abstract}
Natural languages are complexly structured entities. They exhibit characterising regularities that can be exploited to link them one another. In this work, I compare two morphological aspects of languages: \textit{Written Patterns} and \textit{Sentence Structure}. I show how languages spontaneously group by similarity in both analyses and derive an average language distance. Finally, exploiting \textit{Sentence Structure} I developed an Artificial Neural Network capable of distinguishing languages suggesting that not only word roots but also grammatical sentence structure is a characterising trait which alone suffice to identify them.
\end{abstract}

\section{Introduction}
The fact that languages evolved branching from common ancestors is a well established concept\cite{har}. There are some regularities and changes that occur in languages which characterise how they are written and how sentences are formed. Centuries of evolution, migrations and influences among people lead to unique marks which can be used to analyse language similarities and differences. Comparative studies are used to reconstruct the phylogeny of languages and to trace back their origins\cite{LanCla}. 

In recent years, the explosion of open data has lead to an unprecedented proliferation of analyses\cite{rkit} in fields such as personalised medicine\cite{CIRILLO2019161}, to improve buildings energy\cite{BURAKGUNAY201996}, the Internet of Things\cite{URREHMAN2019247} and much more. Data Science and Analytics can now be used in any field to explore new ideas and to support hypotheses. Data Science and Analytics are changing the way we perceive the world. Not only new data generated by social networks are revealing patterns \cite{sana}, but also data as old as our languages still represent a treasure chest to be opened. Languages contain what I would like to call linguistic fossils which had been petrified by aeons of written and oral strata.

In this work, using a computational approach, I compare two morphological aspects of languages: \textit{Written Patterns} and \textit{Sentence Structure}. \textit{Written Patterns} are simple to grasp as they are the result of common roots, for instance, Latin and Greek roots in many European Languages\cite{smi}. \textit{Sentence Structure}, on the other hand, is a more subtle feature to catch but some patterns do occur with regularity. I combined these two analyses to derive a language similarity tree which takes into account these two aspects. 

Several strategies to identify languages have been proposed in the past with good results\cite{compare} \cite{compare2} \cite{compare3} but, to my knowledge, no approach based exclusively on part of speech and Artificial Neural Networks has been explored. As \textit{Sentence Structure} analysis proved to be a good language classifier, I trained an Artificial Neural Network to prove that it is possible to recognise a language exclusively from the way sentences a build, neglecting the words themselves thus showing that \textit{Sentence Structure} is a language specific trait.

\section{Results}

I started my analysis with two exploratory steps: \textit{Written Patters Analysis} and \textit{Sentence Structure Analysis}. By combining these two analyses I derived an \textit{Overall Similarity} of the languages at study. In both my exploratory analyses I took distinctive elements for each language and calculated their relative frequencies. To compare languages I calculated pair-wise similarities.

In every step of this exploration, languages spontaneously group together in their families and/or groups. The large Indo-European family is clearly distinguishable in every analysis. Romance Languages grouped together, Germanic languages, Uralic languages, Baltic languages, Slavic languages and others. Among other results, the exploratory data analysis revealed a similarity between Turkish and Basque.

Following data exploration, I investigated the possibility of identifying languages from the sentence structure alone. \textit{Sentence Structure Analysis} indicated that structures among similar languages do differ, thus suggesting they can be used to identify languages without taking words into consideration. To this end, I trained an Artificial Neural Network that can recognise languages only looking at how sentences are structured with an accuracy of 96.85\% (ten time cross validated with standard deviation 1.47\%).

\subsection{Written Patterns Analysis}

I analysed 10,000 phrases for each language transliterating them into Latin alphabet and sub-setting them in units of two and three characters.\\
\\
\begin{figure}[htp]
\centering
\includegraphics[scale=0.70]{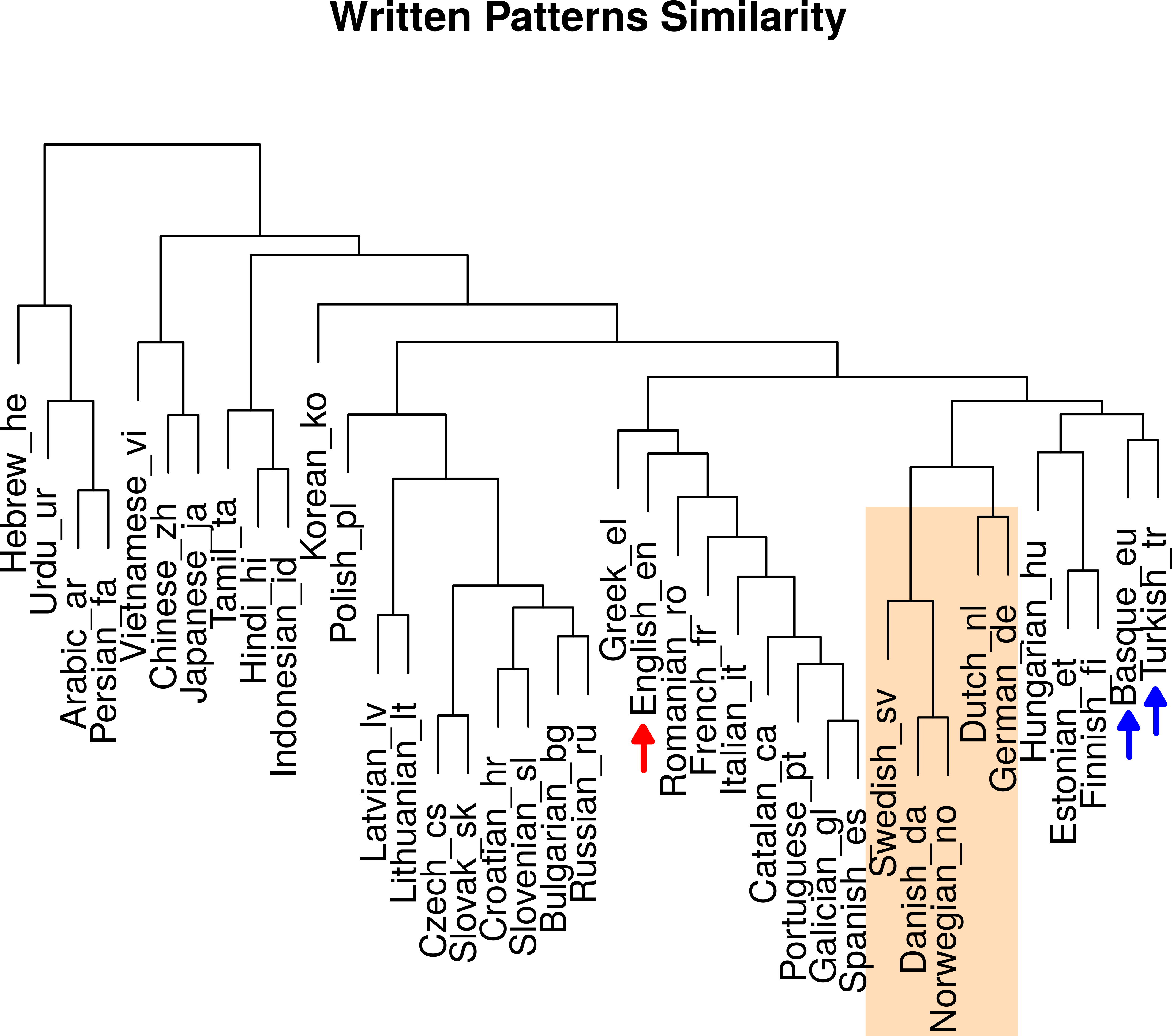}
\caption{Languages tend to group together according to how similar their spellings are. The orange area highlights Germanic Languages where English (red arrow), whose spelling was influenced by Latin and French, is not included. Blue arrows show the interesting closeness of the Turkish language to the Basque Language.}
\label{Figure 1}
\end{figure}
\\
Example:\\
The word "WORD" is broken down into the following\\
tri-grams: \{WOR\}, \{ORD\} \\
di-grams: \{WO\}, \{OR\}, \{RD\}\\
\\
The first observation for \textit{Written Patterns Analysis} is that the Indo-European group (influenced by Latin and Greek) is clearly distinguishable, this can be seen as a positive control; an expected result that had to be true (Figure \ref{Figure 1}).

Germanic languages form a cluster with the interesting absence of the English language, which is better classified under the Romance Languages group due to the important influence Latin first, and French later, had on it\cite{smi}\cite{wien}, making it diverge from Germanic languages (Figure \ref{Figure 1}). Italian for instance, is closer to French than it is to Spanish, and for linguists this is not a surprise - it is another positive control.

\begin{table}
\begin{adjustwidth}{-1.6in}{-2in} 
\scriptsize
\begin{tabular}{|c|c|c|c|c|c|}
\hline
\makecell{\textbf{Arabic}\\al, an, ha, la, dh\\dha, ala, hdh, aan, alm} & 
\makecell{\textbf{Basque}\\en, ar, er, te, an\\tze, zen, eta, egi, bat} & 
\makecell{\textbf{Bulgarian}\\na, da, se, va, to\\ata, ite, ova, ava, tov} & 
\makecell{\textbf{Catalan}\\es, en, er, qu, ar\\que, per, est, ent, sta} & 
\makecell{\textbf{Chinese}\\an, ng, sh, en, ia\\ang, ian, shi, hen, xia} & 
\makecell{\textbf{Croatian}\\je, st, ti, ne, li\\ije, sto, sam, ati, sta}\\
\hline
\makecell{\textbf{Czech}\\ne, te, se, st, na\\ste, sem, pro, jse, sta} & 
\makecell{\textbf{Danish}\\er, de, en, et, or\\jeg, det, kke, der, for} & 
\makecell{\textbf{Dutch}\\en, er, et, ee, de\\aar, een, het, iet, nie} & 
\makecell{\textbf{English}\\th, he, ou, in, er\\the, you, ing, hat, and} & 
\makecell{\textbf{Estonian}\\le, se, ta, ma, te\\ole, lle, kui, sel, sin} & 
\makecell{\textbf{Finnish}\\ta, an, in, aa, en\\tta, sta, aan, lla, taa} \\
\hline
\makecell{\textbf{French}\\ou, es, re, le, en\\ous, vou, que, ais, est} & 
\makecell{\textbf{Galician}\\es, de, on, en, os\\que, non, est, ent, sta} & 
\makecell{\textbf{German}\\ch, en, er, ie, ic\\ich, ein, cht, sie, sch} & 
\makecell{\textbf{Greek}\\ei, to, ou, te, na\\eis, tha, nai, the, ein} & 
\makecell{\textbf{Hebrew}\\wt, ny, ym, hy, hw\\hyy, ywt, hyh, kwl, shw} & 
\makecell{\textbf{Hindi}\\ha, ra, ai, ar, ka\\ara, hai, aha, aim, ata} \\
\hline
\makecell{\textbf{Hungarian}\\el, gy, en, eg, sz\\nem, meg, egy, sze, ogy} & 
\makecell{\textbf{Indonesian}\\an, ka, ng, ak, da\\ang, kan, men, aku, nya} & 
\makecell{\textbf{Italian}\\er, re, on, no, to\\che, non, per, ent, are} & 
\makecell{\textbf{Japanese}\\an, sh, hi, in, ng\\shi, ian, ang, hit, ing} & 
\makecell{\textbf{Korean}\\eo, eu, an, ul, on\\eul, eon, eun, yeo, neu} & 
\makecell{\textbf{Latvian}\\ie, es, ka, as, ta\\vin, ies, man, ina, par} \\
\hline
\makecell{\textbf{Lithuanian}\\ai, as, ka, is, ta\\tai, man, iau, kad, pas} & 
\makecell{\textbf{Norwegian}\\er, en, de, et, eg\\jeg, det, kke, ikk, for} & 
\makecell{\textbf{Persian}\\sh, kh, my, ay, wn\\khw, ayn, ash, awn, dar} & 
\makecell{\textbf{Polish}\\ie, ni, ze, na, sz\\nie, dzi, wie, sie, rze} & 
\makecell{\textbf{Portuguese}\\es, ar, qu, de, ra\\que, nao, est, sta, com} & 
\makecell{\textbf{Romanian}\\in, ca, re, ar, at\\est, are, ine, int, ste} \\
\hline
\makecell{\textbf{Russian}\\to, na, ne, et, st\\cto, eto, ego, ost, pro} & 
\makecell{\textbf{Slovak}\\ne, to, st, na, ie\\som, nie, pre, sta, pri} & 
\makecell{\textbf{Slovenian}\\je, se, ne, po, na\\kaj, sem, pri, pre, bil} & 
\makecell{\textbf{Spanish}\\es, en, de, er, ue\\que, est, ent, sta, ien} & 
\makecell{\textbf{Swedish}\\ar, an, de, en, er\\jag, det, att, for, har} & 
\makecell{\textbf{Tamil}\\ka, an, al, na, tu\\kal, nka, atu, nan, lla} \\
\hline
\makecell{\textbf{Turkish}\\in, ir, ar, an, en\\bir, yor, ini, sin, iyo} & 
\makecell{\textbf{Urdu*}\\y?, my, ?y, ??, ky\\?y?, my?, n?y, ayk, awr} & 
\makecell{\textbf{Vietnamese}\\ng, oi, on, an, nh\\ong, toi, hon, ung, kho} &
&
& \\
\hline
\end{tabular}
\end{adjustwidth}
\caption{Most frequent two-characters and three-character sequences for each language.\newline
* Some characters in the Urdu language were not converted into Latin alphabet} \label{t1}
\end{table}

It is interesting to notice how some languages came together as expected. Finnish, Hungarian and Estonian are the languages spoken in three not adjacent countries which are related\cite{fihues} and known to share common features such as agglutination. The presence of the Turkish language can make this group also be interpreted as the debated Ural-Altaic language family which is currently only a speculation. An interesting result in this group is the Basque language, whose origins are yet to be clarifies - it is not a new theory that Basque and Turkish may be related\cite{batur}. Finally, Korean, an isolated language, is actually  dangling on its own between Asian languages and European ones.

As an overview of each language, Table \ref{t1} lists the 5 most frequent 2-character-elements and 3-character-elements for each analysed language.

\subsection{Sentence Structure Analysis}

To perform this analysis I tagged each word in each sentence with its role in the phrase. Part-of-speech tags were grouped in sets of three elements like I did for the \textit{Written Patterns Analysis}.\\
\\
Example:\\
\begin{tabular}{|c|c|c|c|c|c|}
\hline
	auxiliary & pronoun & adposition & determiner & noun & punctuation\\
\hline
	Are & you & on & this & boat & ?\\
\hline
\end{tabular}
\\
\\
Is broken down into the following tri-grams:\\
\{AUX, PRON, ADP\}, \{PRON, ADP, DET\}, \{ADP, DET, NOUN\}, \{DET, NOUN, PUNCT\} \\

\begin{figure}[htp]
\centering
\includegraphics[scale=0.70]{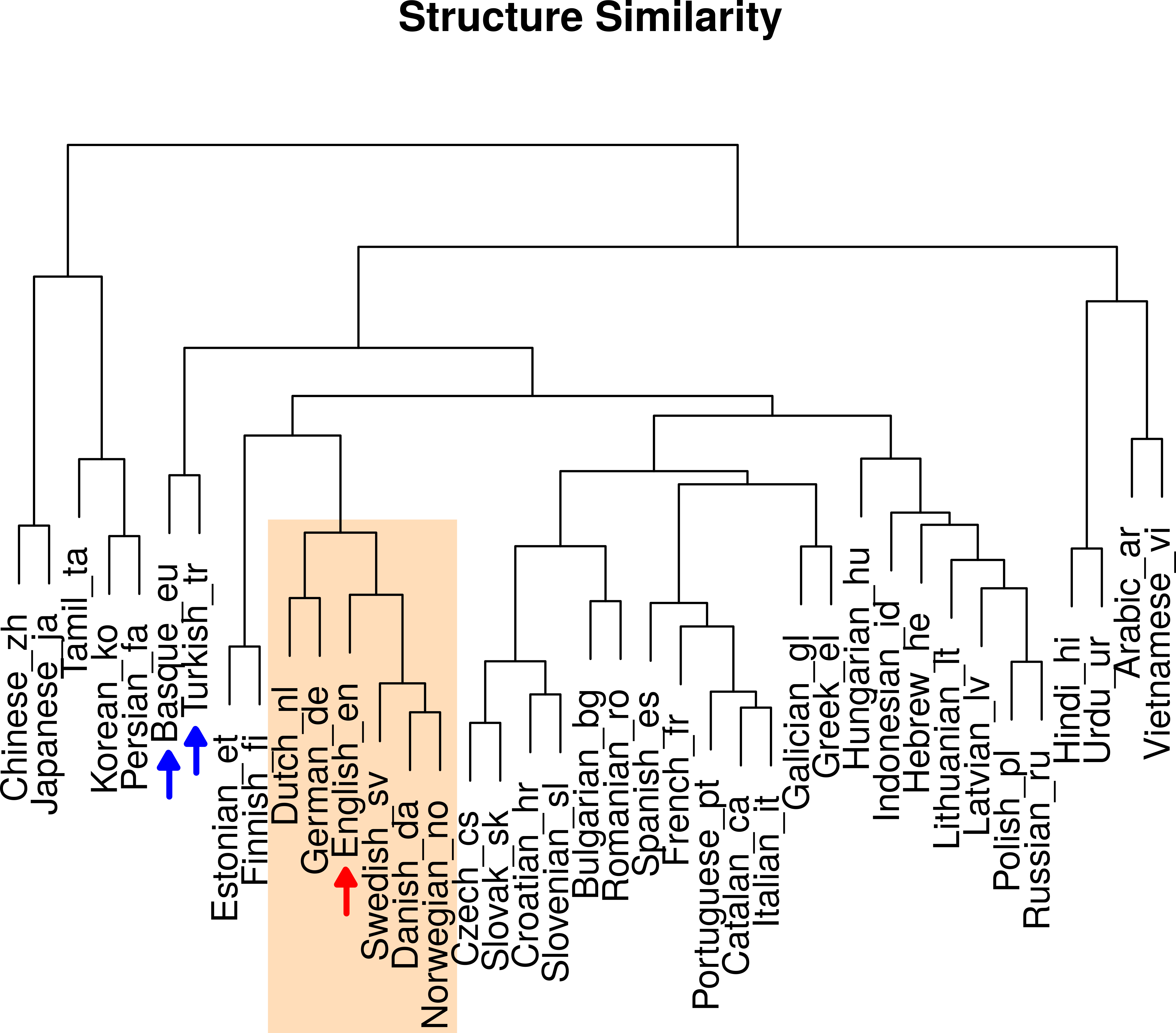}
\caption{Languages tend to group together according to how similar their structures are. Differently from Written Patterns Similarity, English structure make it group with Germanic languages (red arrow). Also in this case Basque and Turkish are close together.}
\label{Figure 2}
\end{figure}

\textit{Structure Structure Analysis} is surely harder to comprehend and possibly it does not mean that distant languages are related. On the other hand, languages like Romance Languages are expected to share similar sentence structures due to their common origin. 

\begin{table}
\begin{adjustwidth}{-1.6in}{-2in} 
\scriptsize
\begin{tabular}{|c|c|c|c|c|c|}
\hline
\makecell{\textbf{ Arabic ar }\\VERB ADP NOUN\\PART VERB NOUN\\CCONJ VERB NOUN\\NOUN PART VERB\\ADJ ADP NOUN}&
\makecell{\textbf{ Basque eu }\\PRON NOUN VERB\\NOUN DET VERB\\DET NOUN VERB\\CCONJ NOUN VERB\\NOUN ADV VERB}&
\makecell{\textbf{ Bulgarian bg }\\VERB ADP NOUN\\PRON VERB ADP\\ADP ADJ NOUN\\ADP NOUN PRON\\VERB NOUN ADP}&
\makecell{\textbf{ Catalan ca }\\ADP DET NOUN\\VERB DET NOUN\\DET NOUN ADP\\NOUN ADP DET\\VERB ADP DET}&
\makecell{\textbf{ Chinese zh }\\ADJ NOUN ADV\\NOUN PROPN ADV\\ADJ ADV NOUN\\PROPN ADV NOUN\\ADV PROPN NOUN}&
\makecell{\textbf{ Croatian hr }\\VERB ADP NOUN\\ADP DET NOUN\\ADP ADJ NOUN\\VERB DET NOUN\\PRON VERB ADP}\\
\hline
\makecell{\textbf{ Czech cs }\\VERB ADP NOUN\\ADP DET NOUN\\ADP ADJ NOUN\\VERB DET NOUN\\DET ADJ NOUN}&
\makecell{\textbf{ Danish da }\\ADP DET NOUN\\VERB PRON ADV\\DET ADJ NOUN\\PRON VERB ADV\\ADV VERB PRON}&
\makecell{\textbf{ Dutch nl }\\ADP DET NOUN\\DET NOUN ADP\\VERB PRON ADV\\DET ADJ NOUN\\PRON VERB ADV}&
\makecell{\textbf{ English en }\\ADP DET NOUN\\DET NOUN ADP\\VERB DET NOUN\\DET ADJ NOUN\\NOUN ADP DET}&
\makecell{\textbf{ Estonian et }\\VERB PRON NOUN\\PRON NOUN VERB\\PRON VERB NOUN\\PRON ADV VERB\\ADV PRON VERB}&
\makecell{\textbf{ Finnish fi }\\VERB PRON NOUN\\PRON VERB NOUN\\VERB PRON ADV\\VERB ADV NOUN\\VERB ADJ NOUN}\\
\hline
\makecell{\textbf{ French fr }\\ADP DET NOUN\\VERB DET NOUN\\DET NOUN ADP\\NOUN ADP DET\\VERB ADP DET}&
\makecell{\textbf{ Galician gl }\\VERB DET NOUN\\DET NOUN ADP\\DET NOUN VERB\\PRON VERB DET\\DET NOUN ADJ}&
\makecell{\textbf{ German de }\\ADP DET NOUN\\DET NOUN VERB\\DET ADJ NOUN\\DET NOUN ADP\\PRON VERB ADV}&
\makecell{\textbf{ Greek el }\\VERB DET NOUN\\DET NOUN PRON\\ADP DET NOUN\\PART VERB DET\\PRON DET NOUN}&
\makecell{\textbf{ Hebrew he }\\VERB ADP NOUN\\PRON VERB NOUN\\PRON ADV VERB\\VERB PART NOUN\\ADV VERB NOUN}&
\makecell{\textbf{ Hindi hi }\\PRON NOUN ADP\\ADP NOUN VERB\\PRON NOUN VERB\\NOUN ADP VERB\\DET NOUN ADP}\\
\hline
\makecell{\textbf{ Hungarian hu }\\VERB DET NOUN\\DET ADJ NOUN\\ADV DET NOUN\\PRON DET NOUN\\DET NOUN VERB}&
\makecell{\textbf{ Indonesian id }\\PRON ADV VERB\\VERB ADP NOUN\\ADV VERB NOUN\\PRON VERB NOUN\\PROPN ADV VERB}&
\makecell{\textbf{ Italian it }\\VERB DET NOUN\\ADP DET NOUN\\DET NOUN ADP\\DET ADJ NOUN\\NOUN ADP DET}&
\makecell{\textbf{ Japanese ja }\\NOUN PART ADV\\NOUN PART ADJ\\PART ADV NOUN\\NOUN PART PROPN\\NOUN PART CCONJ}&
\makecell{\textbf{ Korean ko }\\NOUN ADV VERB\\ADV VERB NOUN\\ADV NOUN VERB\\VERB NOUN ADV\\NOUN VERB ADV}&
\makecell{\textbf{ Latvian lv }\\PRON VERB NOUN\\PRON VERB ADV\\SCONJ PRON VERB\\VERB ADP NOUN\\PRON ADV VERB}\\
\hline
\makecell{\textbf{ Lithuanian lt }\\PRON VERB NOUN\\VERB ADP NOUN\\PRON NOUN VERB\\ADP DET NOUN\\VERB DET NOUN}&
\makecell{\textbf{ Norwegian-Bokmaal no }\\VERB PRON ADP\\PRON VERB ADP\\PRON VERB ADV\\ADP DET NOUN\\DET ADJ NOUN}&
\makecell{\textbf{ Persian fa }\\ADP PRON NOUN\\PRON NOUN VERB\\NOUN ADP PRON\\ADP NOUN VERB\\PRON ADP NOUN}&
\makecell{\textbf{ Polish pl }\\VERB ADP NOUN\\VERB PRON ADP\\PRON ADP NOUN\\ADP DET NOUN\\ADP ADJ NOUN}&
\makecell{\textbf{ Portuguese pt }\\VERB DET NOUN\\ADP DET NOUN\\DET NOUN ADP\\DET NOUN ADV\\DET NOUN VERB}&
\makecell{\textbf{ Romanian ro }\\VERB ADP NOUN\\ADP NOUN DET\\DET NOUN ADP\\PART PRON VERB\\NOUN ADP PRON}\\
\hline
\makecell{\textbf{ Russian ru }\\VERB ADP NOUN\\ADP ADJ NOUN\\PRON PART VERB\\ADP DET NOUN\\VERB ADP PRON}&
\makecell{\textbf{ Slovak sk }\\VERB ADP NOUN\\ADP DET NOUN\\SCONJ PRON VERB\\VERB DET NOUN\\ADP ADJ NOUN}&
\makecell{\textbf{ Slovenian sl }\\VERB ADP NOUN\\ADP DET NOUN\\ADP ADJ NOUN\\VERB DET NOUN\\SCONJ PRON VERB}&
\makecell{\textbf{ Spanish es }\\ADP DET NOUN\\VERB DET NOUN\\DET NOUN ADP\\NOUN ADP DET\\VERB ADP DET}&
\makecell{\textbf{ Swedish sv }\\PRON VERB ADV\\ADP DET NOUN\\ADV VERB PRON\\DET ADJ NOUN\\VERB PRON ADV}&
\makecell{\textbf{ Tamil ta }\\PRON ADJ NOUN\\PRON DET NOUN\\PRON NOUN ADJ\\ADV PRON NOUN\\PRON ADV NOUN}\\
\hline
\makecell{\textbf{ Turkish tr }\\PRON NOUN VERB\\ADJ NOUN VERB\\NOUN ADJ VERB\\ADV NOUN VERB\\DET NOUN VERB}&
\makecell{\textbf{ Urdu ur }\\ADP NOUN VERB\\PRON ADP NOUN\\PRON NOUN ADP\\NOUN PRON ADP\\NOUN ADP VERB}&
\makecell{\textbf{ Vietnamese vi }\\PROPN VERB NOUN\\VERB ADP NOUN\\VERB NOUN PROPN\\VERB NOUN ADP\\NOUN VERB PROPN}&
&
& \\
\hline
\end{tabular}
\end{adjustwidth}
\caption{Most frequent grammatical three-elements-structures consisting of three different part-of-speech elements used for each language.} \label{t4}
\end{table}

Also in this second analysis I calculated language distances and drew a tree shown in Figure \ref{Figure 2}. In this second case though, groups are not intuitive and not easy to comment. I noticed some similarities which confirm \textit{Written Pattern Analysis} and some groups that are in contrast with it.

Romance Languages are still grouped together, Slavic languages, Uralic languages and others. A good result is the one of the English language whose spelling moved it into the Romance Languages group but its structure moved it to group together with Germanic Languages, where it historically belongs. Italian also in this case, tends to stay closer to French than it is to Spanish.

Another interesting result is, again, the Basque language, which falls close to Turkish\cite{batur}, with which it might share some feature derived by agglutination. Turkish and Basque similarity is debatable but both my analyses confirm it.

Other similarities such as Korean, Persian and Tamil are harder to comment and need further analyses. Nevertheless, these groups might just be due to POS tagging limits and errors. Table \ref{t4} shows most frequent grammatical sentence structures for every language.

\begin{figure}[htp]
\centering
\includegraphics[scale=0.55]{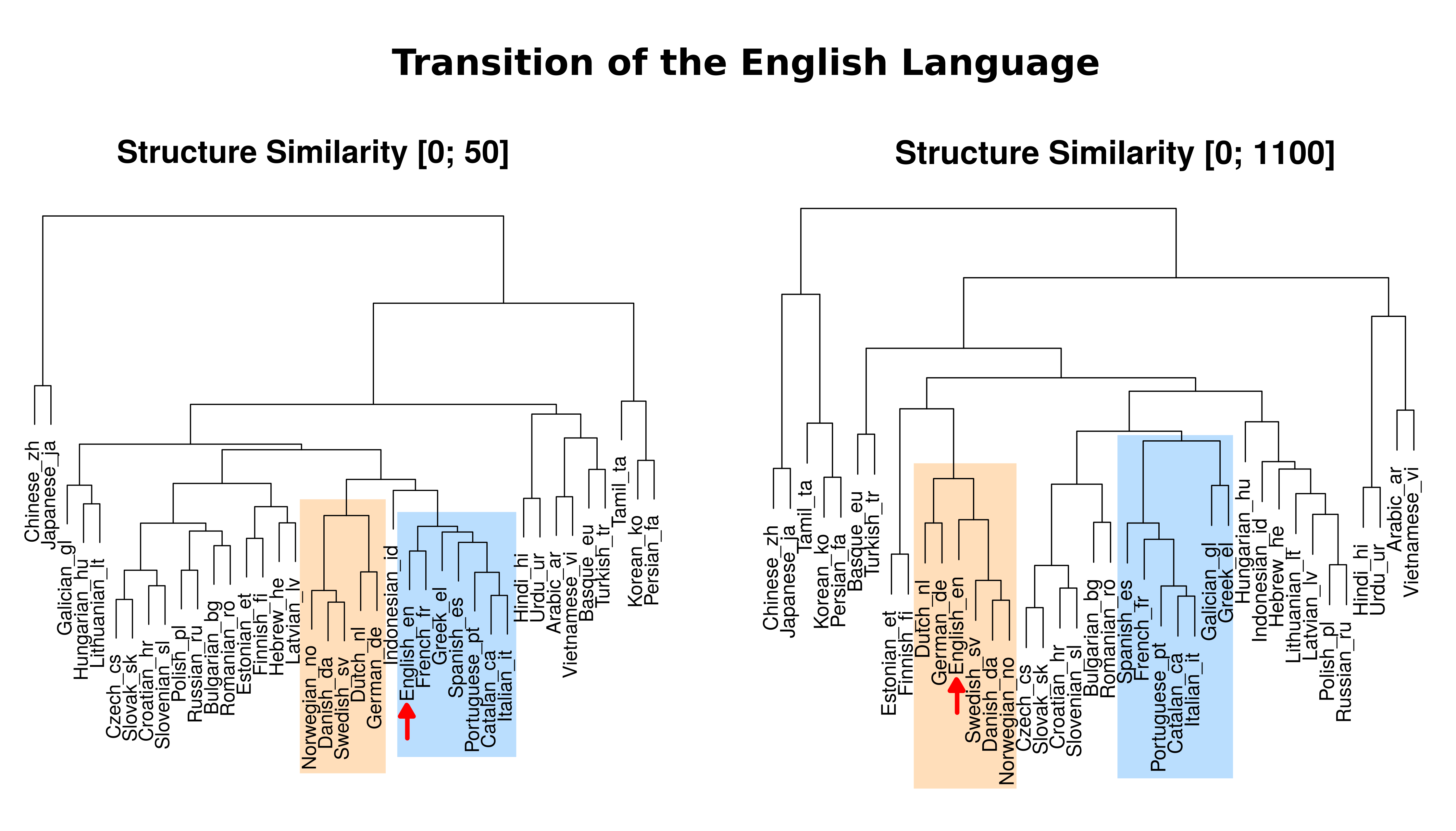}
\caption{The blue area shows Romance Languages with Greek. The orange area shows Germanic languages. The Red arrow shows the transition of English from Romance languages to Germanic languages as more sentence structure features are considered.}
\label{figure eng}
\end{figure}

To further illustrate that sentence structure hides meaningful language features, I tried to cluster languages considering top 50 and top 1100 sentence structures of three elements. Figure \ref{figure eng} shows how the large Romance and Germanic languages are already clustered with just 50 features. Furthermore, it is interesting to notice how the most frequent sentence structures of the English Language place it in the Romance languages while least frequent structures - possibly more articulated sentences - move the English Language within the Germanic Language family.

Even though this analysis is hard to comment, it does show some regularities that can rise the question if such differences might suffice to identify languages. 

\subsection{Overall Similarity}

\begin{figure}[htp]
\centering
\includegraphics[scale=0.70]{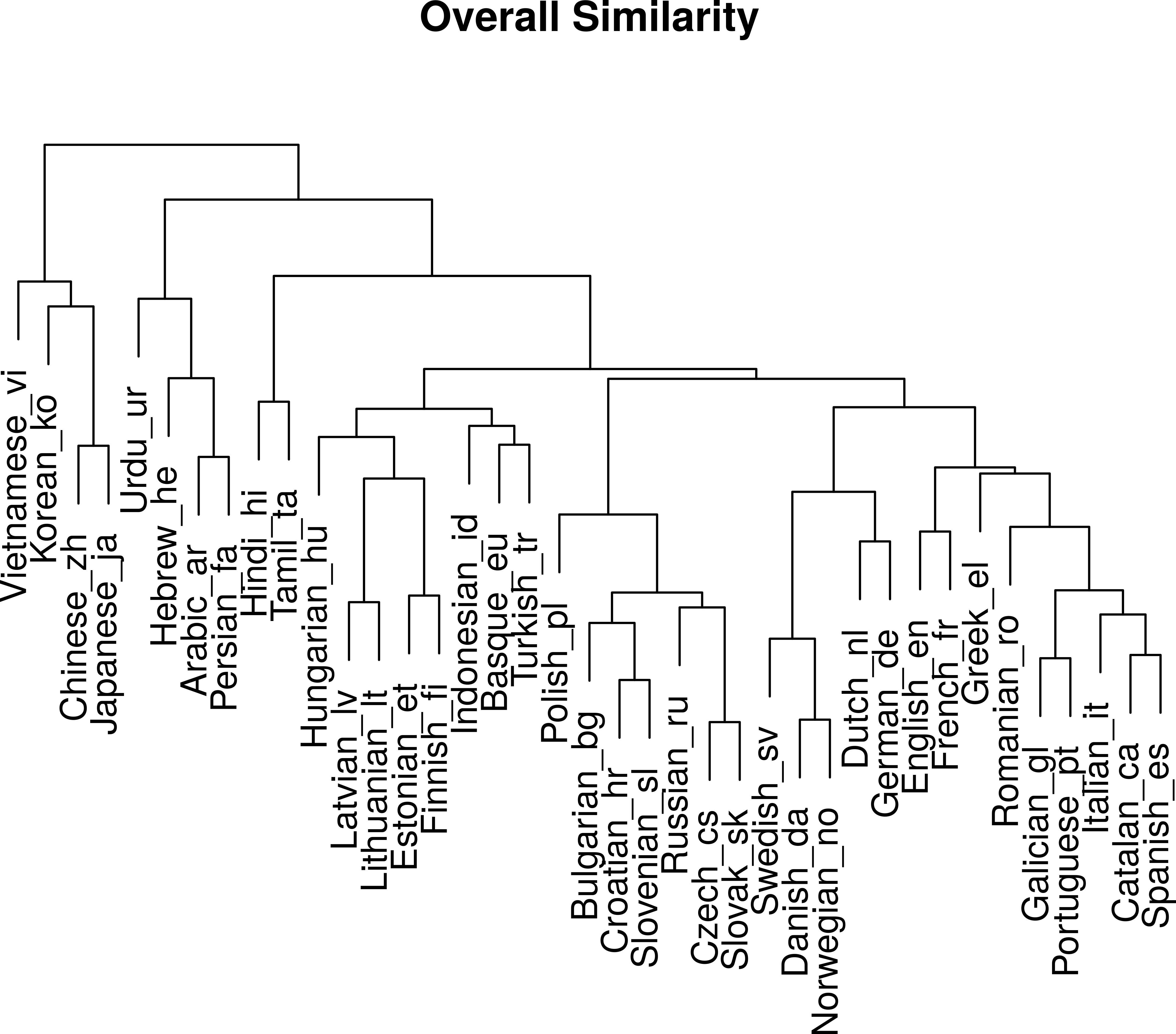}
\caption{Languages tend to group together according to both Written and Sentence Structure similarities.}
\label{Figure 3}
\end{figure}

I averaged the two distances obtained in \textit{Written Pattern Analysis} and \textit{Sentence Structure Analysis} to derive an overall similarity. This \textit{Overall Similarity} is not intended to be a language classification but maybe it shows how easy it would be for a speaker of one language to learn another language and, maybe most importantly, how easy it would be to actually sound natural when speaking and writing it. In this exploratory analysis I only comment on the Italian language, my first language. I can confirm that despite grammatical and written similarities with French, the overall effort to learn it might actually be more than it is to learn Spanish. The tree in Figure \ref{Figure 3} shows the average similarity among the languages at study.

\begin{figure}[htp]
\centering
\includegraphics[scale=0.70]{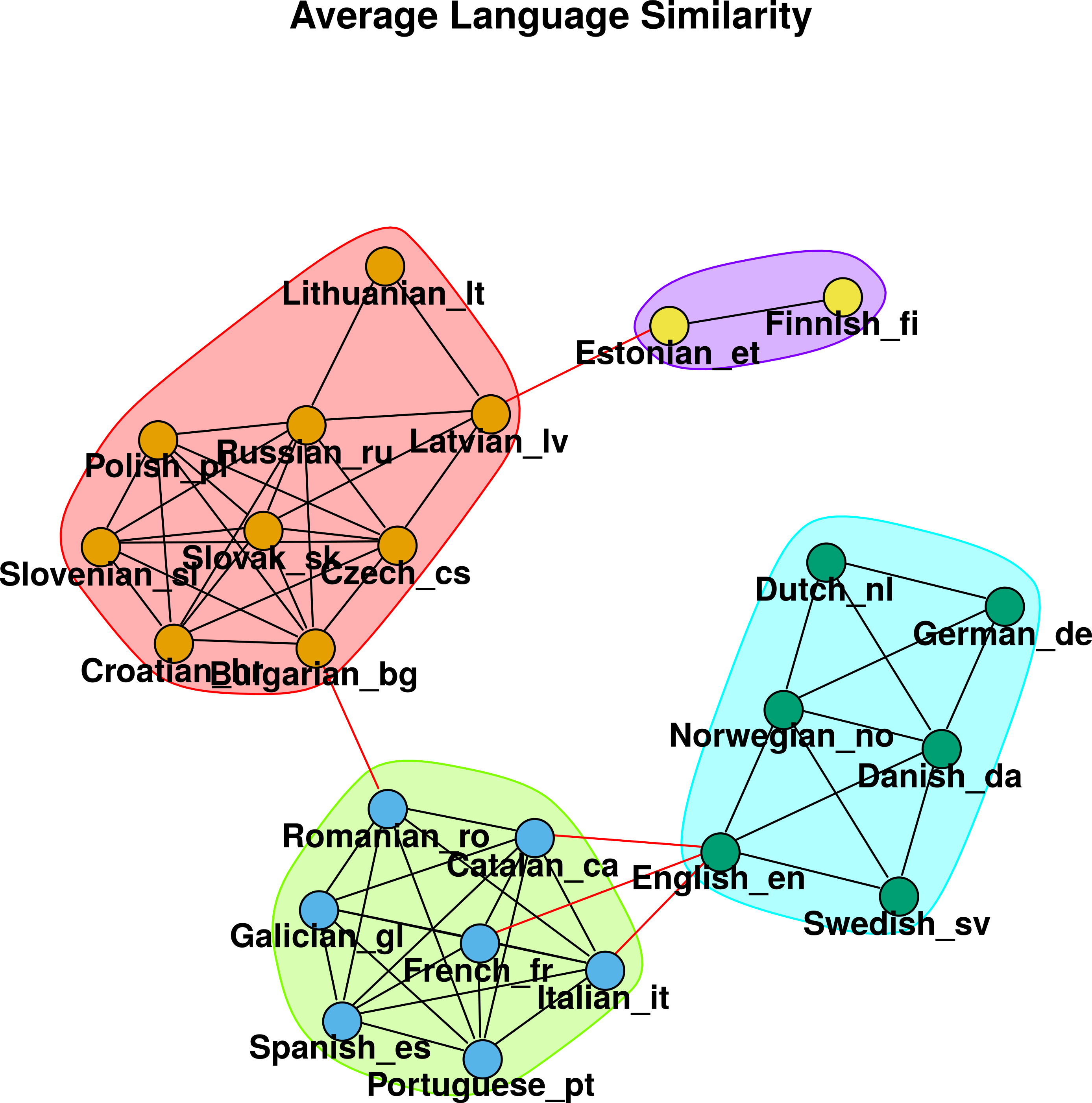}
\caption{Languages with a similarity with a confidence greater than 75\%. Cyan are Germanic Languages, green Romance Languages, red are Slavic Languages and purple Uralic languages. Red edges are the bridges connecting two groups.}
\label{Figure 4}
\end{figure}

A second analysis I conducted exploits graph theory. I took the same average similarities used to plot the tree and filtered them to keep only similarities with a significance greater than 75\% (I used z-score 1.15035). Plotting the remaining pairs I obtained a graph (Figure \ref{Figure 4}) whose calculated clusters are: Romance Languages, Germanic Languages, Slavic Languages and Uralic Languages. It is interesting to notice the "bridges" among these clusters. The graph could be read like "If I speak Italian, learning English could help me stepping into Germanic Languages" or also "If I speak Italian, learning Romanian could help me stepping into Slavic Languages" or "If I speak a Romance Language, Uralic Languages can be a challenge".

\subsection{Sentence Structure Aware Artificial Neural Network}
While comparing words spelling and roots is a common way to discriminate languages I am not aware of any study which demonstrates the order of parts of speech is a language specific trait.
\\
\\
Can one tell which languages are passages 1 and 2?
\begin{enumerate}
\item particle verb particle verb determiner noun punctuation adverb punctuation noun verb punctuation verb determiner noun noun noun noun punctuation subordinating conjunction particle verb particle determiner verb adposition noun adposition adjective noun punctuation
\item adverb pronoun coordinating conjunction verb noun coordinating conjunction adjective punctuation adverb pronoun auxiliary noun adverb adposition noun punctuation pronoun verb adverb pronoun verb adposition adverb adposition particle adjective pronoun adposition noun noun punctuation
\end{enumerate}

It turns out an Artificial Neural Network can.\\
\\
Taking into account the results obtained in \textit{Structure Structure Analysis} I built an Artificial Neural Network which takes as input the probability of three-elements part of speech in a sentence and tries to guess the source language.
The Artificial Neural Network I developed was able to recognise languages only from sentence structure with a good accuracy of 96.85\% with a standard deviation of 1.47\% (ten fold cross validated) which demonstrates that the order of parts of speech in a sentence is just as specific as the number of languages used in this analysis. Table \ref{tnn} shows precision, recall and F-score for each language. Performances are always good with the exception of some languages like Czech and Slovak whose structure might be confused as they are probably cognate languages.

\begin{table}
\scriptsize
\begin{center}
\begin{tabular}{|c|c|c|c|}
\hline
&\textbf{Precision}&\textbf{Recall}&\textbf{F-score}\\
\hline
Arabic&0,99&1&1\\
\hline
Basque&1&1&1\\
\hline
Bulgarian&0,96&0,98&0,97\\
\hline
Catalan&0,96&0,98&0,97\\
\hline
Chinese&1&1&1\\
\hline
Croatian&0,74&0,62&0,67\\
\hline
Czech&0,81&0,9&0,85\\
\hline
Danish&0,96&0,93&0,94\\
\hline
Dutch&0,99&0,98&0,99\\
\hline
English&1&1&1\\
\hline
Estonian&0,98&0,96&0,97\\
\hline
Finnish&0,96&0,98&0,97\\
\hline
French&0,97&0,99&0,98\\
\hline
Galician&1&1&1\\
\hline
German&0,98&0,99&0,99\\
\hline
Greek&0,99&1&1\\
\hline
Hebrew&1&1&1\\
\hline
Hindi&0,99&1&0,99\\
\hline
Hungarian&1&1&1\\
\hline
Indonesian&1&0,99&0,99\\
\hline
Italian&0,98&0,96&0,97\\
\hline
Japanese&1&1&1\\
\hline
Korean&1&1&1\\
\hline
Latvian&0,99&1&1\\
\hline
Lithuanian&0,99&0,99&0,99\\
\hline
Norwegian&0,95&0,98&0,96\\
\hline
Persian&1&1&1\\
\hline
Polish&0,95&0,92&0,93\\
\hline
Portuguese&0,97&0,97&0,97\\
\hline
Romanian&0,99&0,98&0,99\\
\hline
Russian&0,94&0,96&0,95\\
\hline
Slovak&0,82&0,76&0,79\\
\hline
Slovenian&0,78&0,86&0,82\\
\hline
Spanish&0,99&0,97&0,98\\
\hline
Swedish&0,96&0,99&0,97\\
\hline
Tamil&1&1&1\\
\hline
Turkish&1&0,98&0,99\\
\hline
Urdu&0,98&0,95&0,97\\
\hline
Vietnamese&1&1&1\\
\hline
\end{tabular}\caption{Artificial Neural Network performance for each languages. Some languages are predicted with an F-Score of 1, some others, like Czech and Slovak which are some what harder to recognise due to the fact that they are cognate languages} \label{tnn}
\end{center}
\end{table}

The language of passage 1 is Greek and the language of passage 2 is Swedish. Respectively: "Θα προσπαθήσουμε να κατακτήσουμε τον κόσμο. Επιτέλους, καθηγητά Μιφούνε, τελειοποιήσατε τη διαδικασία συρρίκνωσης ηλεκτρονικών συσκευών, ώστε να μπορέσουμε να την πουλήσουμε στους αμερικάνους για τρελά χρήματα." and "Där det varken fanns smärta eller lidande. Där det var skratt istället för död. Jag trodde alltid hon hittade på allt för att trösta mig under stundens smärta".

\section{Materials and Methods}

\subsection{Materials}
The entire computational analysis was conducted using the R programming language\cite{rprog}.

The corpora used for this analysis were downloaded from OPUS\cite{tide}. I downloaded OpenSubtitles v2016 (http://www.opensubtitles.org/). Languages used for this analysis are those present in both OpenSubtitles v2016 and RDRPOSTagger. I took and preprocessed (see Methods) 100,000 lines for each language.

For part-of-speech tagging I used the RDRPOSTagger package\cite{pos}. To identify languages I used the "Google's Compact Language Detector 3" package. For sentence tokenization I used the package tokenizers\cite{tok}. To transliterate languages I used the function stri\textunderscore trans\textunderscore general in the package stringi\cite{{stri}}.

A graph was built calculation the similarity z-score and filtering only connections with 75\% confidence - i.e. z-score=1.15035. To analyse the language graph I used igraph\cite{igraph}. In particular, communities\cite{fg} were detected with Infomap algorithm\cite{Rosvall1118}.

To develop the Artificial Neural Network I used the python programming language\cite{pprog}. I used Keras\cite{chollet2015keras} with TensorFlow\cite{tensorflow2015-whitepaper} backend, SciKit Learn\cite{scikit-learn} for cross validation and Pandas\cite{pandas} to manipulate large datasets.

\subsection{Methods}
To avoid mixed-language sentences in the corpora at study, texts were preprocessed to be sure that each phrase was actually in the target language and not containing spurious words and characters. Secondly, sentences shorter than 3 words were excluded. Finally, possible duplicated sentences were removed.

\begin{table}
\begin{adjustwidth}{-1in}{-1in} 
\scriptsize
\begin{center}
\begin{tabular}{|c|c|c|c|}
\hline
\makecell{\textbf{Language and code}} & \makecell{\textbf{Clean sentences}} & \makecell{\textbf{Number of features}\\Written Patterns Analysis}& \makecell{\textbf{Number of features}\\Sentence Structure Analysis}\\
\hline
Arabic ar & 27,924 & 2,000 & 2,000\\
\hline
Basque eu & 46,176 & 2,000 & 2,000\\
\hline
Bulgarian bg & 52,102 & 2,000 & 2,000\\
\hline
Catalan ca & 47,370 & 2,000 & 2,000\\
\hline
Chinese zh & 10,594 & 2,000 & 363\\
\hline
Croatian hr & 20,412 & 2,000 & 2,000\\
\hline
Czech cs & 43,662 & 2,000 & 2,000\\
\hline
Danish da & 37,864 & 2,000 & 2,000\\
\hline
Dutch nl & 45,518 & 2,000 & 2,000\\
\hline
English en & 48,742 & 2,000 & 2,000\\
\hline
Estonian et & 46,107 & 2,000 & 2,000\\
\hline
Finnish fi & 39,416 & 2,000 & 1,775\\
\hline
French fr & 47,356 & 2,000 & 2,000\\
\hline
Galician gl & 44,031 & 2,000 & 2,000\\
\hline
German de & 49,045 & 2,000 & 2,000\\
\hline
Greek el & 50,821 & 2,000 & 2,000\\
\hline
Hebrew he & 60,004 & 2,000 & 2,000\\
\hline
Hindi hi & 29,111 & 2,000 & 2,000\\
\hline
Hungarian hu & 44,259 & 2,000 & 2,000\\
\hline
Indonesian id & 30,884 & 2,000 & 2,000\\
\hline
Italian it & 45,837 & 2,000 & 2,000\\
\hline
Japanese ja & 12,287 & 2,000 & 727\\
\hline
Korean ko & 23,850 & 2,000 & 775\\
\hline
Latvian lv & 43,057 & 2,000 & 2,000\\
\hline
Lithuanian lt & 37,068 & 2,000 & 2,000\\
\hline
Norwegian no & 43,486 & 2,000 & 2,000\\
\hline
Persian fa & 28,821 & 2,000 & 2,000\\
\hline
Polish pl & 45,778 & 2,000 & 2,000\\
\hline
Portuguese pt & 42,995 & 2,000 & 2,000\\
\hline
Romanian ro & 38,009 & 2,000 & 2,000\\
\hline
Russian ru & 42,492 & 2,000 & 2,000\\
\hline
Slovak sk & 41,835 & 2,000 & 2,000\\
\hline
Slovenian sl & 39,985 & 2,000 & 2,000\\
\hline
Spanish es & 39,794 & 2,000 & 2,000\\
\hline
Swedish sv & 43,806 & 2,000 & 2,000\\
\hline
Tamil ta & 9,755 & 2,000 & 1,259\\
\hline
Turkish tr & 40,811 & 2,000 & 1,761\\
\hline
Urdu ur & 7,127 & 2,000 & 2,000\\
\hline
Vietnamese vi & 61,805 & 2,000 & 2,000\\
\hline
\end{tabular}
\end{center}
\end{adjustwidth}\caption{List of languages analysed with their language codes, number of phrases remaining after preprocessing and number of features (di-grams and tri-grams) considered for each language for written and structural analyses. Some languages have fewer POS-tri-grams than others just because of their nature, for example agglutinations (Finnish and Turkish) and the scarcity or lack of inflections and abundance of particles(Chinese and Japanese)} \label{t2}
\end{table}

To perform \textit{Written Patterns Analysis} I divided texts in di-grams and tri-grams (units consisting of two and three characters respectively), excluding spaces. For each n-gram, I calculated its probability dividing the number of its occurrences by the total number of n-grams. I took the top-1000 di-grams and the top-1000 tri-grams for each language, for a total of 2000 features for each language.
\\
\\
\\
\\
\\
Example:\\
\\
The word "WORD" is broken down into the following\\
tri-grams: \{WOR\}, \{ORD\} \\
di-grams: \{WO\}, \{OR\}, \{RD\}\\
\\
To account for different alphabets I transliterated languages into diacritics-free Latin alphabet. This allows a direct comparison of languages overcoming the alphabet barrier. Thought it might not be always accurate, it grants that diacritics (accents) on some letters do not result in different n-grams. For instance, the Greek word "πρόβλημα", which means \textit{problem} in English has no accent over the letter o but it is clearly the same word. Table \ref{t2} summarises sentences and features used for each language.
\\
\\
Example:\\
\\
The sentence: Το τηλέφωνό μου έχει πρόβλημα\\
In Latin with diacritics: To tēléphōnó mou échei próblēma\\
Latin without diacritics: To telephono mou echei problema\\
We can easily spot the word "telephone" and the word "problem". Plus, the word "mou" is very similar to "my".
\\
\\
To perform \textit{Sentence Structure Analysis}, I first converted phrases into part-of-speech (POS) elements and then I built POS-tri-grams. Also in this case, I considered 2,000 features for each language, if available.
\\
\\
Example:\\
\\
\begin{tabular}{|c|c|c|c|c|c|}
\hline
	auxiliary & pronoun & adposition & determiner & noun & punctuation\\
\hline
	Are & you & on & this & boat & ?\\
\hline
\end{tabular}
\\
\\
Is broken down into the following tri-grams:\\
\{AUX, PRON, ADP\}, \{PRON, ADP, DET\}, \{ADP, DET, NOUN\}, \{DET, NOUN, PUNCT\} 
\\

\begin{table}
\scriptsize
\begin{center}
\begin{tabular}{|c|c|}
\hline
	\textbf{Tag} & \textbf{Part of speech}\\
\hline
	ADJ & adjective\\
\hline
	ADP & adposition\\
\hline
	ADV & adverb\\
\hline
	AUX & auxiliary\\
\hline
	CCONJ & coordinating conjunction\\
\hline
	DET & determiner\\
\hline
	INTJ & interjection\\
\hline
	NOUN & noun\\
\hline
	NUM & numeric\\
\hline
	PART & particle\\
\hline
	PRON & pronoun\\
\hline
	PROPN & proper noun\\
\hline
	PUNCT & punctuation\\
\hline
	SCONJ & subordinating conjunction\\
\hline
	SYM & symbol\\
\hline
	VERB & verb\\
\hline
	X & other\\
\hline
\end{tabular}
\end{center}\caption{List of part of speech considered for sentence structure analysis.} \label{t3}
\end{table}

Table \ref{t3} lists all grammatical elements considered for Sentence Structure Analysis. These tags were selected as they are the subset of tags derivable by all languages compared in this analysis conducted using the UniversalPOS tagger (see Materials).

For both analysis, each language was represented as a vector of many components and, due to the high dimensionality of these vectors, the Manhattan distance was used to calculate distances\cite{man}.
\\

For the \textit{Overall Similarity} I averaged the \textit{Written Patterns Analysis} and the \textit{Sentence Structure Analysis}. To plot the similarity graph I calculated the z-score of these averaged values and filtered out values outside a confidence interval of 75\% (z-score 1.15035).
{}
\\

To train the Artificial Neural Network I randomly extracted 100 sentences for each language and combined them into single documents. I generated 1,000 documents for each language and collected them into one single dataset. For each document I calculated the probability of POS-tri-grams excluding those containing the element X. Other than the input and output layers, the neural network consisted of two hidden layers:
\begin{itemize}
\item Layer 1: 2018 inputs and 8 outputs. Relu activation function\cite{relu}
\item Layer 2: 8 inputs and 39 outputs. Softmax activation function which is good to classify mutually exclusive classes as one class (ground-truth) gets score 1 while other labels will get 0. \cite{Goodfellow-et-al-2016} 
\end{itemize}

I used categorical cross-entropy\cite{cross} as loss function. Adam\cite{kingma2014method} as optimiser. Finally, I used a 10 times cross validation to validated the model. 

\section{Conclusions and Further Developments}
Language classification has always been conducted by comparative approaches. Computational methods allow a thorough analysis which automatically does exactly the same work but faster and against a very large number of words and sentences. With the aid of automatic calculation I explored morphological language features from a written and structural perspectives.

I commented some known facts like the written similarity of the English languages with Romance Languages and its structural similarity with Germanic languages. This analysis also supports unclear hypothesis like the relationship between the Turkish and Basque languages. Analysing two morphological aspects of many languages I speculated on the easiness of learning a foreign language.

Finally, I developed an Artificial Neural Network which can recognise languages only from the order of part of speech demonstrating words order is a language specific trait just like vocabulary and pronunciation.

The study presented in this work can be applied to more languages. It can be further developed to analyse more features including pronunciation using the International Phonetic Alphabet. It can be generalised to analyse different writing styles: scientific, journalistic, narrative and maybe, by means of a well-trained Artificial Neural Network, even identify the real native language of a person writing in a second language. Finally, but maybe most importantly, analysing languages with corpora written in the past, can be used to trace evolutionary changes in sentence structures of human languages. 

\bibliography{MyBib.bib}
\bibliographystyle{ieeetr}
\end{document}